\documentclass[10pt,twocolumn,letterpaper]{article}

\usepackage[pagenumbers]{cvpr} 

\definecolor{cvprblue}{rgb}{0.21,0.49,0.74}
\usepackage[pagebackref,breaklinks,colorlinks,allcolors=cvprblue]{hyperref}

\usepackage[utf8]{inputenc} 
\usepackage[T1]{fontenc}    
\usepackage{url}            
\usepackage{booktabs}       
\usepackage{amsfonts}    
\usepackage{nicefrac}    
\usepackage{microtype}      
\usepackage{xcolor}         
\usepackage{graphicx}
\usepackage{multirow}

\def\etal{\emph{et al}}

\usepackage{pifont}
\usepackage{enumitem}

\def\confName{CVPR}
\def\confYear{2025}

\title{OpenStereo: A Comprehensive Benchmark for Stereo Matching \\and Strong Baseline}

\author{
\vspace{-5mm}
 \\
  Xianda Guo$^{1}$~~~~~~~
  Chenming Zhang$^{2*}$~~~~~~~
  Juntao Lu$^{2*}$~~~~~~~
  Yiqun Duan$^{2}$, \\
  Yiqi Wang$^{2}$~~~~~~~
  Tian Yang$^{3}$~~~~~~~ 
  Zheng Zhu$^{3}$~~~~~~~ 
  Long Chen$^{2,4,\dag}$ \\
  \textsuperscript{1} School of Computer Science, Wuhan University~~~~\textsuperscript{2}Waytous~~~~\textsuperscript{3} GigaAI\\
  \textsuperscript{4} Institute of Automation, Chinese Academy of Sciences
\\
\url{https://github.com/XiandaGuo/OpenStereo}\\
\texttt{xianda\_guo@163.com}
\vspace{-5mm}
}

\begin{document}

\maketitle

\renewcommand{\thefootnote}{\fnsymbol{footnote}}
\footnotetext[1]{Joint second authors;
\footnotemark[2] Corresponding author}

\begin{abstract}
Stereo matching aims to estimate the disparity between matching pixels in a stereo image pair, which is important to robotics, autonomous driving, and other computer vision tasks. Despite the development of numerous impressive methods in recent years, determining the most suitable architecture for practical application remains challenging. To address this gap, our paper introduces a comprehensive benchmark focusing on practical applicability rather than solely on individual models for optimized performance. Specifically, we develop a flexible and efficient stereo matching codebase, called \textbf{OpenStereo}. OpenStereo includes training and inference codes of more than 10 network models, making it, to our knowledge, the most complete stereo matching toolbox available. Based on OpenStereo, we conducted experiments and have achieved or surpassed the performance metrics reported in the original paper.
Additionally, we conduct an exhaustive analysis and deconstruction of recent developments in stereo matching through comprehensive ablative experiments. These investigations inspired the creation of \textbf{StereoBase}, a strong baseline model. Our StereoBase ranks 1\textsuperscript{st} on SceneFlow, KITTI 2015, 2012 (Reflective) among published methods and performs best across all metrics. In addition, StereoBase has strong cross-dataset generalization. 

\end{abstract}
\section{Introduction}
\label{sec:intro}


Stereo matching is a fundamental topic in the field of computer vision, aiming to compute the disparity between a pair of rectified stereo images. It plays a crucial role in numerous applications such as robotics~\cite{linestereo2015}, autonomous driving~\cite{orbslam22017, shamsafar2022mobilestereonet}, and augmented reality~\cite{smart_iotj2020}, as it enables depth perception and 3D reconstruction of the observed scene.

\begin{figure*}[t]
\begin{center}
 \includegraphics[width=1\linewidth]{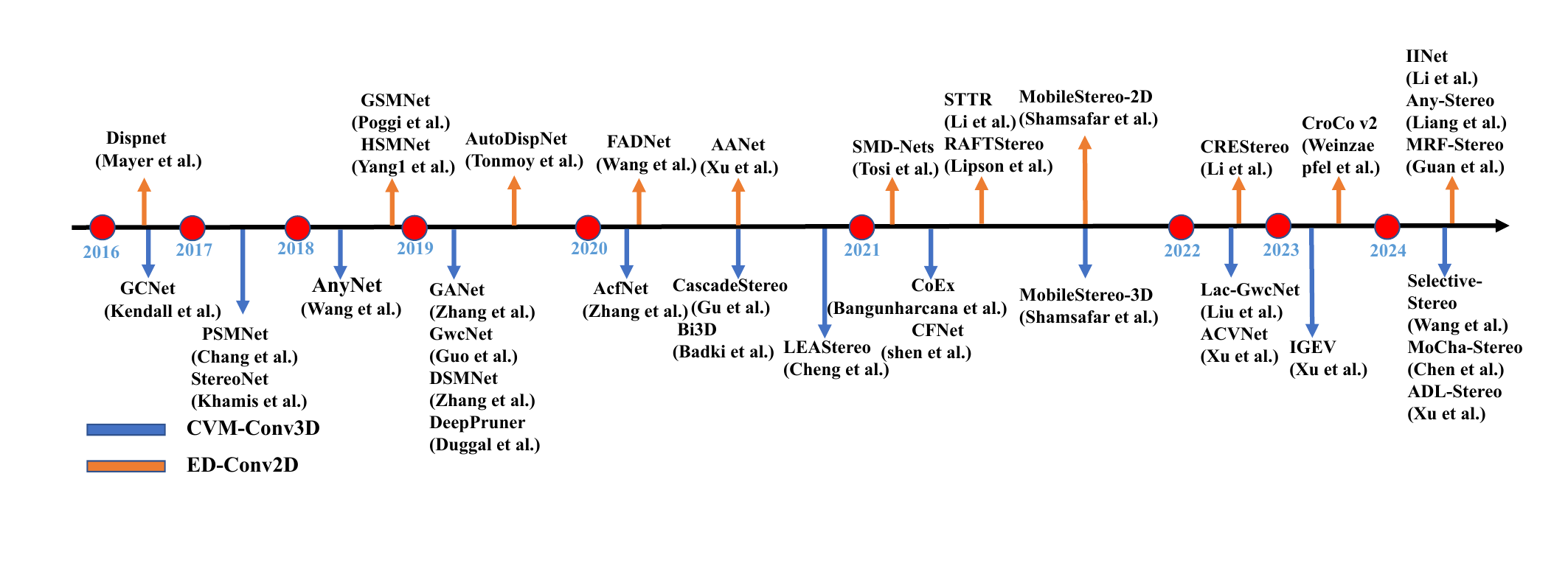}
\end{center}
\vspace{-1.0cm}
\caption{\textbf{Timeline of Stereo Matching Models.} The top part shows ED-conv2D-based models, while the bottom part shows CVM-conv3D-based models. Each model is labeled with its name and authors.}
\label{fig:timeline}
\end{figure*}

Traditional stereo-matching algorithms typically match corresponding image regions between the left and right views based on their similarity measures. Several techniques have been proposed in the literature for stereo matching, including methods based on gray-level information~\cite{Birchfield1999APDM, Li2013EDS, Yang2010ASA}, region-based approaches~\cite{Zhang2005SPS, Pinggera2015EGS}, and energy optimization methods~\cite{Scharstein2002ATaE,hirschmuller2007sgm}. 
With the support of large synthetic datasets~\cite{sceneflow, middlebury, eth3d,kitti2012, kitti2015, yang2019drivingstereo}, CNN-based stereo matching methods~\cite{gcnet, psmnet2018, gwcnet2019, xu2023iterative} has achieved impressive results. 
As shown in \autoref{fig:timeline}, based on the network pipeline of stereo matching, CNN-based stereo matching methods can be roughly grouped into two categories~\cite{wang2021fadnet++}, including the encoder-decoder network with 2D convolution (ED-Conv2D)~\cite{sceneflow,gsmnet,hsmnet2019,AutoDispNet,wang2021fadnet++,xu2020aanet,tosi2021smd,STTR,raftstereo,Crestereo,croco_v2,li2024iinet} and the cost volume matching with 3D convolution (CVM-Conv3D)~\cite{gcnet,psmnet2018,ganet2019,gwcnet2019,dsmnet2020,deeppruner2019,zhang2019adaptive,gu2020cascade,badki2020bi3d,leastereo,bangunharcana2021coex,cfnet2021,acvnet,xu2023iterative,chen2024mocha,ganetADL}. 

However, we find that different studies often employ various data augmentation strategies, learning rates, learning rate optimization methods, and backbone architectures. This inconsistency makes it difficult to evaluate and compare various methods' performance accurately. 
This inconsistency in experimental setups and methodologies makes it difficult to derive conclusive insights and hampers the objective assessment of advancements in stereo matching. Without a standardized benchmark, researchers struggle to identify the true impact of individual components and innovations. There is a lack of clear conclusions and exploration regarding data augmentation strategies, backbone selection, and cost construction methods in stereo matching.

Moreover, not all datasets are accompanied by official evaluation tools. 
The SceneFlow~\cite{sceneflow} dataset, with its finalpass and cleanpass data varieties, complicates fair model comparisons. Generalization experiments for stereo matching algorithms typically train on the SceneFlow dataset and evaluate on KITTI2012~\cite{kitti2012}, KITTI2015~\cite{kitti2015}, ETH3D~\cite{eth3d}, and Middlebury~\cite{middlebury}. 
However, due to the absence of a standard protocol for generalization experiments, different papers may yield inconsistent results for the same method. For instance, discussions on the generalization performance of IGEV~\cite{xu2023iterative} across works~\cite{xu2023iterative,wang2024selective,guan2024neural} exemplify this issue.

Hence, there's a pressing need for a comprehensive benchmark study within the stereo-matching community to enhance practicality and ensure consistent comparisons. To achieve this objective, we introduce a versatile stereo-matching codebase: OpenStereo.
To promote scalability and adaptability, OpenStereo offers the following features: (1) 
\textbf{Modulal design}, researchers can define a new model without the need to alter the model code itself by simply modifying a YAML configuration file. (2)\textbf{Various frameworks}, including Concatenation-based~\cite{psmnet2018,ganet2019}, Correlation-based~\cite{xu2020aanet, wang2020fadnet, wang2021fadnet++,bangunharcana2021coex}, Interlaced-based~\cite{shamsafar2022mobilestereonet}, Group-wise-correlation-based~\cite{xu2023iterative}, Combine-based methods~\cite{gwcnet2019}, and Difference-based~\cite{stereonet2018}. (3)\textbf{Various datasets}, including SceneFlow~\cite{sceneflow}, KITTI2012~\cite{kitti2012}, KITTI2015~\cite{kitti2015}, Middlebury~\cite{middlebury}, and ETH3D~\cite{eth3d}. (4) \textbf{State-of-the-art methods}, including PSMNet~\cite{psmnet2018}, GwcNet~\cite{gwcnet2019}, AANet~\cite{xu2020aanet}, FADNet++~\cite{wang2021fadnet++}, CFNet~\cite{cfnet2021}, STTR~\cite{STTR}, CoEx~\cite{bangunharcana2021coex}, CascadeStereo~\cite{gu2020cascade}, MobileStereoNet~\cite{shamsafar2022mobilestereonet} and IGEV~\cite{xu2023iterative}.

Leveraging OpenStereo, we rigorously reassess various officially stated conclusions by re-implementing the ablation studies, including data augmentation, backbone architectures, cost construction, disparity regression, and refinement processes. Based on the insights gleaned from these ablation experiments, we introduce StereoBase, a model that sets a new benchmark, surpassing recently proposed methods in terms of performance. StereoBase is powerful and serves as an empirically state-of-the-art (SOTA) baseline model for stereo matching, demonstrating exceptional efficacy and resilience across diverse testing scenarios.
Our contributions are summarized as follows:
\begin{itemize}[leftmargin=0.5cm]
\item[$\bullet$] We introduce \textbf{OpenStereo}, a unified and extensible platform, which enables researchers to conduct comprehensive stereo matching studies. 
\item[$\bullet$] We evaluate the impact of CNN, Transformer, and hybrid CNN-Transformer backbones on stereo matching. To our knowledge, this is the first work to explore these architectures.
\item[$\bullet$] We conduct a profound revisitation and thorough deconstruction of recent stereo-matching methodologies. 
\item[$\bullet$] We introduce \textbf{StereoBase}, which sets a new benchmark with EPE of 0.34 on SceneFlow~\cite{sceneflow} and ranks 1\textsuperscript{st} on KITTI2015~\cite{kitti2015} and 2012(Reflective)\cite{kitti2012} leaderboards among published methods. 
\end{itemize}

\section{Related Work}

\subsection{Stereo Matching}
With the rapid development of CNNs, significant progress has been made in stereo matching. Based on the network pipeline of stereo matching,  stereo matching methods can be roughly grouped into two categories~\cite{wang2021fadnet++}, including the encoder-decoder network with 2D convolution (ED-Conv2D) and the cost volume matching with 3D convolution (CVM-Conv3D).

\begin{figure*}[t]
      \centering
      \includegraphics[width=1\linewidth]{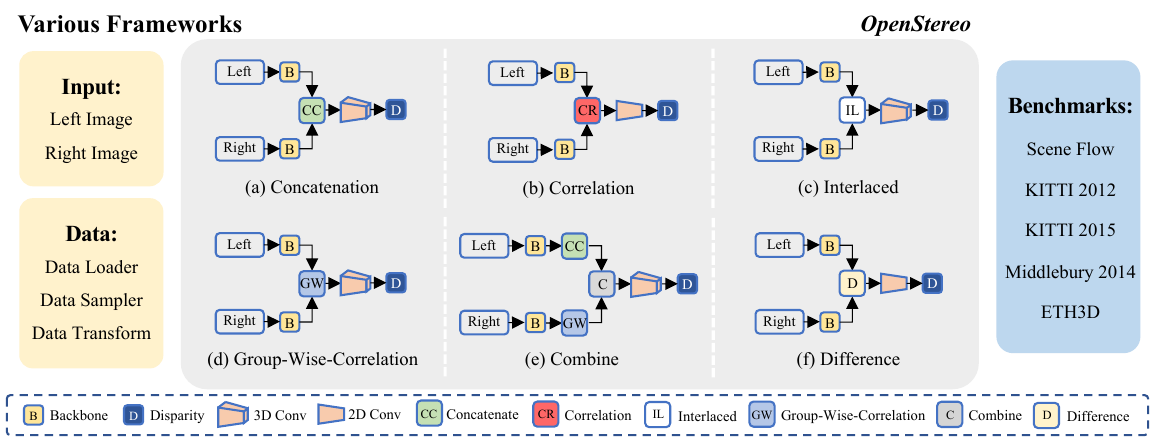}
      \caption{\textbf{The design principles of proposed codebase OpenStereo.}} 
      \label{fig:framework}
      \vspace{-6mm}
\end{figure*}

\noindent
\textbf{Stereo Matching with CVM-Conv3D} 
The CVM-Conv3D methods are proposed to improve the performance of depth estimation~\cite{gcnet,psmnet2018,segstereo2018,anynet2019,ganet2019,gwcnet2019,dsmnet2020,deeppruner2019,zhang2019adaptive,gu2020cascade,badki2020bi3d,leastereo,bangunharcana2021coex,cfnet2021,acvnet,xu2023iterative,ganetADL}. These methods learn disparities from a 4D cost volume, mainly constructed by concatenating left feature maps with their corresponding right counterparts across each disparity level~\cite{psmnet2018}. GCNet~\cite{gcnet} firstly introduced a novel approach that combines 3D encoder-decoder architecture with a 2D convolutional network to obtain a dense feature representation, which is used to regularize a 4D concatenation volume. Following GCNet~\cite{gcnet}, PSMNet~\cite{psmnet2018} proposes an approach for regularizing the concatenation volume by leveraging a stacked hourglass 3D convolutional neural network in tandem with intermediate supervision. To enhance the expressiveness of the cost volume and ultimately improve performance in ambiguous regions, GwcNet~\cite{ganet2019} proposes the group-wise correlation volume and ACVNet~\cite{acvnet} proposes the attention concatenation volume. 
CoEx~\cite{bangunharcana2021coex} proposes a novel approach called Guided Cost volume Excitation (GCE), which leverages image guidance to construct a simple channel excitation of the cost volume. IGEV-Stereo~\cite{xu2023iterative} leverages an iterative geometry encoding volume to capture local and non-local geometry information, outperforming existing methods on KITTI benchmarks and achieving cross-dataset generalization and high inference efficiency. 

However, these CVM-Conv3D methods still suffer from low time efficiency and high memory requirements, which are far from real-time inference even on server GPUs. Therefore, it is essential to address the accuracy and efficiency problems for real-world applications.

\noindent
\textbf{Stereo Matching with ED-Conv2D} The ED-Conv2D methods~\cite{sceneflow,gsmnet,hsmnet2019,AutoDispNet,wang2020fadnet,wang2021fadnet++,xu2020aanet,tosi2021smd,STTR,raftstereo,shamsafar2022mobilestereonet,Crestereo,croco_v2,li2024iinet,guo2024lightstereo}, which adopt networks with 2D convolutions to predict disparity, has been driven by the need for improved accuracy, computational efficiency, and real-time performance. One of the early deep learning-based stereo matching methods, MC-CNN (Matching Cost CNN)~\cite{mccnn}, was proposed to learn a matching cost function for improving performance in the cost aggregation and optimization stages. Subsequently, Mayer \etal ~\cite{sceneflow} present end-to-end networks for the estimation of disparity, called DispNet, which is pure 2D CNN architectures. However, the model still faces challenges in capturing the matching features, resulting in poor estimation results. To overcome this challenge, the correlation layer is introduced in the end-to-end architecture~\cite{sceneflow,flownet,flownet2,flownet3} to better capture the relationship between the left and right images. By incorporating this layer, the accuracy of the model is significantly improved. Furthermore, FADNet++~\cite{wang2021fadnet++} proposes an innovative approach to efficient disparity refinement using residual learning~\cite{resnet2016} in a coarse-to-fine manner. AutoDispNet~\cite{AutoDispNet} applied neural architecture search to automatically design stereo-matching network structures. 
More recently, Croco-Stereo~\cite{croco_v2} shows that large-scale pre-training can be successful for stereo matching through well-adapted pretext tasks. This method can achieve state-of-the-art performance without using task-specific designs, like correlation volume or iterative estimation.

These works represent the significant progress that has been made in the field of stereo matching, highlighting the diverse range of methods and architectures that have been proposed to address the challenges associated with this problem.


\subsection{Codebase}
Numerous infrastructure code platforms have been developed in the deep learning research community, with the aim of facilitating research in specific fields. One such platform is OpenGait~\cite{fan2022opengait}, a gait recognition library. OpenGait thoroughly examines the latest advancements in gait recognition, providing novel perspectives for subsequent research in this domain. In object detection, MMDetection~\cite{mmdetection} and Detectron2~\cite{wu2019detectron2} have emerged as an all-encompassing resource for several favored detection techniques.
In pose estimation, OpenPose~\cite{OpenPose} has developed the first open-source system that operates in real-time for detecting the 2D pose of multiple individuals, including the detection of key points for the body, feet, hands, and face.
In stereo matching, it is noteworthy that not all datasets are accompanied by official evaluation tools. For instance, the DrivingStereo~\cite{yang2019drivingstereo} dataset does not have official evaluation codes, and there is a lack of unified tools for assessing the model generalization across different domains. This absence of standardized evaluation resources contributes to the observed discrepancies in cross-domain evaluations of the same model as reported in different studies. Therefore, it is time to build a benchmark for stereo matching.
\section{OpenStereo}
In recent years, there has been a proliferation of new frameworks and evaluation datasets for stereo-matching. However, the lack of a unified and fair evaluation platform in this field is a significant issue that cannot be ignored. To address this challenge and promote academic research and practical application we have developed \textbf{OpenStereo}, a pyTorch-based~\cite{paszke2019pytorch} toolbox that provides a reliable and standardized evaluation framework for stereo matching.

\subsection{Design Principles of OpenStereo}
As shown in \autoref{fig:framework}, our developed OpenStereo covers the following highlight features.

\noindent
\textbf{Modular Design.} 
OpenStereo adopts a modular design, greatly facilitating researchers in exploring new networks. By simply modifying a YAML configuration file, researchers can define a new model without the need to alter the model code itself. This design significantly lowers the barriers for researchers to extend or integrate additional algorithms and modules within the framework. This approach empowers researchers to freely combine and customize their algorithms with minimal code composition, enhancing the framework's usability and adaptability.

\noindent
\textbf{Compatibility with various frameworks.} 
Currently, more and more stereo matching methods have emerged, such as Concatenation-based~\cite{gcnet,psmnet2018,ganet2019,leastereo}, Correlation-based~\cite{xu2020aanet, wang2020fadnet, wang2021fadnet++,bangunharcana2021coex}, Group-wise-correlation-based~\cite{xu2023iterative}, Difference-based~\cite{stereonet2018}, Interlaced-based~\cite{shamsafar2022mobilestereonet}, and Combine-based methods~\cite{gwcnet2019,cfnet2021}. As mentioned above, many open-source methods have a narrow focus on their specific models, making it challenging to extend to multiple frameworks. However, OpenStereo provides a solution to this problem by supporting all of the aforementioned frameworks. With OpenStereo, researchers and practitioners can easily compare and evaluate different stereo-matching models under a standardized evaluation protocol.

\noindent
\textbf{Support for various evaluation datasets.} 
OpenStereo is a comprehensive tool that not only supports synthetic stereo datasets such as SceneFlow~\cite{sceneflow}, but also four real-world datasets: KITTI2012~\cite{kitti2012}, KITTI2015~\cite{kitti2015}, ETH3D~\cite{eth3d}, and Middlebury~\cite{middlebury}. We introduce a suite of bespoke functions, meticulously crafted for each dataset, encompassing everything from initial data preprocessing to the final stages of evaluation. The evaluation module includes the submission of the results to KITTI2012~\cite{kitti2012} and KITTI2015~\cite{kitti2015} leadboards.

\begin{figure*}[t]
\begin{center}
\vspace{-2mm}
 \includegraphics[width=1\linewidth]{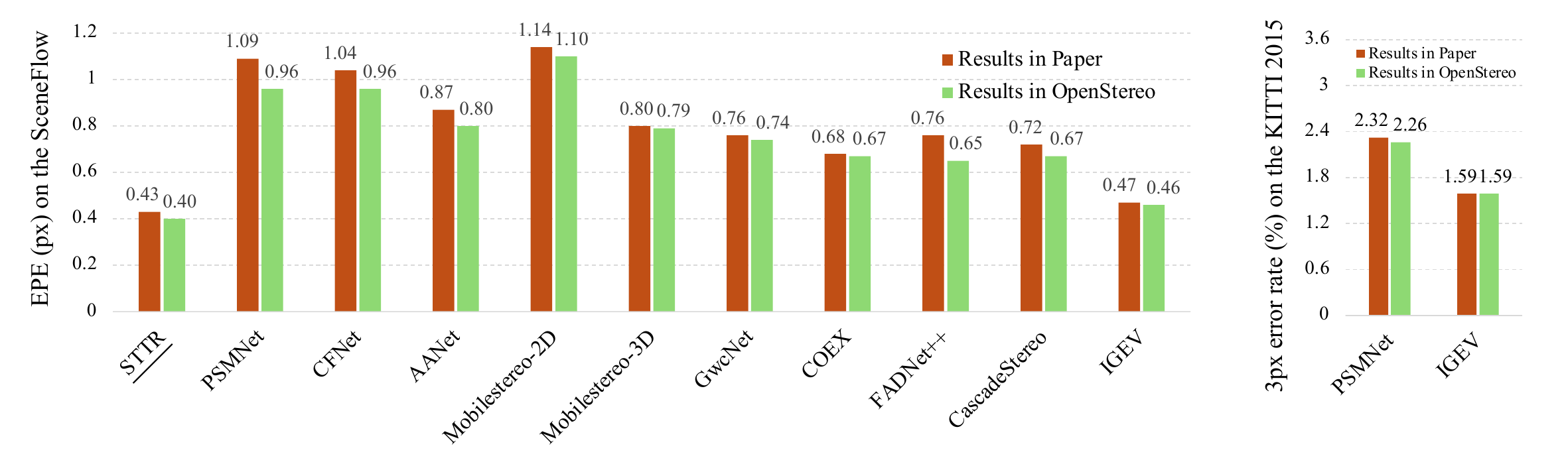}
\end{center}
\vspace{-0.5cm}
\caption{\textbf{Quantitative evaluation on the SceneFlow~\cite{sceneflow} and KITTI2015~\cite{kitti2015} leadboard.} For each model, the specific category on the SceneFlow used is consistent with the original paper. Underline refers to evaluation in the non-occluded regions only STTR~\cite{STTR}.}
\label{fig:bar}
\end{figure*}

\noindent
\textbf{Support for state-of-the-arts.} 
In our work, we have successfully replicated various state-of-the-art stereo matching methods, including PSMNet~\cite{psmnet2018}, GwcNet~\cite{gwcnet2019}, AANet~\cite{xu2020aanet}, FADNet++~\cite{wang2021fadnet++}, CFNet~\cite{cfnet2021}, STTR~\cite{STTR}, CoEx~\cite{bangunharcana2021coex}, CascadeStereo~\cite{gu2020cascade}, MobileStereoNet~\cite{shamsafar2022mobilestereonet} and IGEV~\cite{xu2023iterative}. As shown in \autoref{fig:bar}, the performance metrics we achieved, in most cases, surpass those reported in their original publications.

\section{Revisit Deep Stereo Matching}

\subsection{Datasets and Evaluation Metrics}

\textbf{SceneFlow} \cite{sceneflow} is a synthetic stereo collection, counting 35,454/4,370 image pairs for training/testing, respectively. The data is split into two categories: cleanpass and finalpass. Cleanpass refers to the synthetic images that are generated by clean renderings without post-processing, while finalpass images are produced with photorealistic settings such as motion blur, defocus blur, and noise. The end point error (EPE) is utilized as the evaluation metric.

\textbf{KITTI.} KITTI 2012 \cite{kitti2012} and KITTI 2015 \cite{kitti2015} are datasets that captured from real-world scenes. KITTI 2012 contains 194/195 image pairs for training/testing, while KITTI 2015 provides
200/200 image pairs for training and testing. All KITTI datasets provide sparse ground-truth disparities guided by the LiDAR system. For evaluations, we calculate EPE and the percentage of pixels with EPE larger than 3 pixels in all (D1-all) regions. All two KITTI datasets are also used for cross-domain generalization performance evaluation.

\textbf{Middlebury 2014} \cite{middlebury} has two sets of 15 image pairs for both training and testing, respectively. The stereo images are captured from indoor scenes, with 3 resolution options, which are full, half, and quarter. In evaluations, only the training image pairs with half resolution will be used to evaluate the cross-domain generalization performance, and EPE and 2px Error metric are reported.

\textbf{ETH3D} \cite{eth3d} has stereo images collected from both indoor and outdoor environments with grayscale formate. It comes with 27 and 20 image pairs for training and testing, respectively, and a sparsely labeled ground truth for reference. In evaluation, only the training set is used to report the cross-domain generalization performance. And we adopt the EPE and 1px Error as the metric.


\subsection{Evaluation of Prior Work}

For benchmarking, it is critical to ensure that the results are reliable and trustworthy. To achieve this, we conducted our experiments on SceneFlow~\cite{sceneflow} and KITTI2015~\cite{kitti2015} datasets. Regarding the KITTI2015 dataset, submission constraints led us to limit our leaderboard contributions to reproductions of the widely recognized PSMNet~\cite{psmnet2018} and the latest SOTA IGEV~\cite{xu2023iterative}. As shown in \autoref{fig:bar}, the reproduced performances of OpenStereo are better than the results reported by the original papers. This is primarily due to the normalization applied to the input stereo images by the OpenStereo framework. (All other settings remain consistent with the original paper.)  


\subsection{Necessity of Comprehensive Ablation Study}
In the evolving landscape of stereo-matching, comprehensive ablation studies play a pivotal role in deciphering the effectiveness of different components and strategies. A thorough ablation study goes beyond mere performance metrics; it uncovers the underlying mechanics of different algorithms, revealing their strengths and weaknesses in various scenarios. For instance, different data augmentation techniques may yield contrasting effects on the model's ability to match stereo images accurately. Similarly, the impact of various backbones, cost volume configurations, and disparity regression methods on the overall performance can be profound. Understanding the specific contributions of each component is crucial for building more efficient and effective stereo-matching systems. 

\begin{table}[t]
  \centering
  
   \resizebox{0.48\textwidth}{!}{
    \begin{tabular}{lc|c|cc}
    \toprule
    \multirow{2}{*}{\centering Data Augmentation} & \multirow{2}{*}{\centering LR\_scheduler}&SceneFlow&\multicolumn{2}{c}{KITTI15}  \\ 
      && EPE(px) & EPE(px) & D1\_all \\
    \midrule
    RC(320$\times$736) &MultiStepLR& 0.6839 &2.91&15.73\\
    RC(320$\times$736) &OneCycleLR& 0.6155 &2.34&11.86\\
    \midrule
    RC(256$\times$512) &OneCycleLR&0.6470 &3.02 &14.95\\
    RC(320$\times$736) &OneCycleLR& 0.6155 &2.34&11.86\\
    RC(320$\times$736)+Scale &OneCycleLR& 0.6867&2.88&12.91\\
    RC(320$\times$736)+HFlip &OneCycleLR&0.6612 &2.22&12.27 \\
    RC(320$\times$736)+ColorAug &OneCycleLR& 0.6529&1.68&7.89 \\
    RC(320$\times$736)+VFlip &OneCycleLR& 0.6367&2.09&10.11 \\
    RC(320$\times$736)+Erase &OneCycleLR&0.6167 &2.68&12.64\\
    RC(320$\times$736)+HSFlip &OneCycleLR&\textbf{0.6076} &2.74&13.99\\
    RC(320$\times$736)+CE &OneCycleLR&0.6486 &1.65&8.15\\
    RC(320$\times$736)+CES+HSFlip& OneCycleLR&0.7165 &1.71&8.40\\
    \underline{RC(320$\times$736)+CES} &OneCycleLR& 0.7240&\textbf{1.56}&\textbf{7.64}\\
    \bottomrule
    \end{tabular}%
    }
  \caption{\textbf{Ablation study on Data Augmentation and LR\_scheduler Selection.} KITTI2015~\cite{kitti2015} training set, consisting of 200 images, is only employed to evaluate the generalizability of models. RC stands for RandomCrop~\cite{krizhevsky2012imagenet}. HFlip~\cite{krizhevsky2012imagenet} denotes both images of a stereo and disparity are horizontally flipped. HSFlip~\cite{krizhevsky2012imagenet} horizontally flips both images in the stereo pair and then swaps them. VFlip~\cite{krizhevsky2012imagenet} involves vertically flipping both images in the stereo pair along with the disparity, inverting their top-bottom orientation. CES represents ColorAug~\cite{krizhevsky2012imagenet}, Erase~\cite{zhong2020random}, and Scale~\cite{simonyan2014very}. Settings used in our final model are underlined.} 
  \vspace{-5mm}
  \label{tab:ablation_dataaug}%
\end{table}%

\subsection{Denoising Stereo Matching}

With the support of OpenStereo, a comprehensive reevaluation of various stereo-matching methods is conducted, including data augmentation, feature extraction, cost construction, disparity prediction, and refinement.  Our ablation studies have revealed some new insights. 

\textbf{ImpletionDetails} For the experiments of this subsection, we adopt the COEX~\cite{bangunharcana2021coex} as the baseline model due to its speed. All experiments in this subsection are conducted on SceneFlow~\cite{sceneflow} datasets. FLOPs are calculated at a resolution of $544 \times 960$. The total epochs is set to 50.



\begin{figure*}[t]
  \centering
  \includegraphics[scale=0.78]{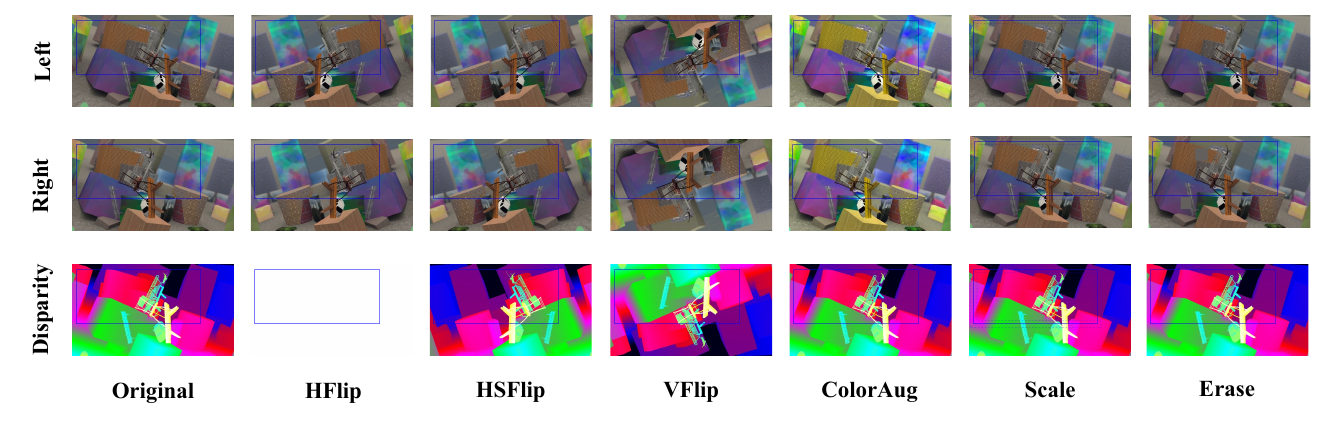}
  \caption{The visualization of stereo images and disparity with different data augmentation. The blue box represents the area where random cropping occurs during training. Notably, when the views are horizontally flipped, the disparity is multiplied by $-1$, making the disparity map appear pure white.} 
  \label{fig:visdataaug}
  \vspace{-0.5cm}
\end{figure*}

\subsubsection{LR\_scheduler and Data Augmentation}

As shown in \autoref{tab:ablation_dataaug}, MultiStepLR yields an EPE of 0.6839, while OneCycleLR achieves a lower EPE of 0.6155. This substantial difference underscores the crucial role of selecting an appropriate learning rate scheduler for stereo matching. 
The superior performance of OneCycleLR indicates its potential to improve model accuracy and robustness, making it a preferable choice over MultiStepLR for training stereo-matching models.
Although five data augmentation techniques—random crop, color augmentation, eraser transform, flip, and spatial transform—are commonly used in stereo matching~\cite{raftstereo, xu2023iterative}, their empirical efficacy specifically for stereo matching has not been thoroughly explored. This study investigated these data augmentation strategies to address this gap. 
Visualizations of these augmentations are provided in \autoref{fig:visdataaug} for further insight. 
Most data augmentation strategies, except for the combined use of HSFlip and random crop, lead to a decline in the model's EPE metric of SceneFlow. This is because stereo matching involves pixel-level matching, and these data augmentations (color augmentation and spatial transform) can affect the alignment of pixels. 
The combination of ColorAug, Erase, and Scale (CES) shows the best generalization performance on KITTI2015, with the lowest EPE of 1.56 and D1\_all of 7.64, although it increases the EPE on SceneFlow to 0.7240.
These findings underscore the importance of selecting appropriate data augmentation methods to enhance model accuracy and robustness.

\begin{table}[t]
  \centering
  
    \resizebox{0.49\textwidth}{!}{
    \begin{tabular}{lcc|ccc}
    \toprule
    Backbone & Type &Pretrain&Flops&Params& EPE \\
    \midrule
    MobilenetV2~\cite{sandler2018mobilenetv2} &CNN&&70.58G&2.78M &0.77 \\
    MobilenetV2~\cite{sandler2018mobilenetv2} &CNN& \checkmark&70.58G&2.78M &\textbf{0.62} \\
    \midrule
    MobilenetV2~\cite{sandler2018mobilenetv2} &CNN&\checkmark&\textbf{70.58G}&\textbf{2.78M} &0.62 \\
    MobilenetV2$^*$~\cite{sandler2018mobilenetv2} &CNN&\checkmark&85.93G&5.21M& 0.56 \\
    EfficientNetV2~\cite{tan2021efficientnetv2} &CNN & \checkmark&157.52G&24.92M&0.52 \\
    RepViT~\cite{wang2023repvit}&Trans.&\checkmark&101.35G&5.64M&0.59\\
    MPViT~\cite{lee2022mpvit} &CNN\&Trans.&\checkmark&283.35G&13.33M &\textbf{0.51}\\
    \bottomrule
    \end{tabular}%
    }
    \caption{\textbf{Ablation study on Backbones Selection}. MobilenetV2$^*$ refers to MobilenetV2 120d. \textbf{Bold}: Best.}
  \label{tab:ablation_backbone}%
  \vspace{-0.5cm}
\end{table}%

\subsubsection{Feature Extraction} 
As shown in \autoref{tab:ablation_backbone}, pretraining the backbone is crucial for stereo matching as it enhances the model's ability to extract robust and informative features.
Furthermore, the choice of backbone significantly influences the model's performance and computational efficiency. 
MobilenetV2~\cite{sandler2018mobilenetv2} and EfficientNetV2~\cite{tan2021efficientnetv2} are lightweight CNNs that are particularly efficient in extracting local features, which are crucial for stereo matching. Their designs allow them to perform well with relatively low computational complexity.
RepViT~\cite{wang2023repvit} is a Transformer-based architecture, which excels in capturing long-range dependencies and global context. While RepViT captures global features well, it might struggle with the fine-grained, pixel-level accuracy required for precise stereo matching.
MPViT~\cite{lee2022mpvit} combines the strengths of both CNNs and Transformers. The CNN components effectively capture local features, while the Transformer components excel in modeling global context. This hybrid approach allows MPViT to leverage the advantages of both architectures, resulting in the lowest EPE. 
In summary, MobilenetV2~\cite{sandler2018mobilenetv2} offers a good balance for applications with limited computational resources, while more complex architectures like EfficientNetV2 and MPViT provide superior accuracy at the cost of higher computational requirements.  
To the best of our knowledge, our work is the first to explore the transformer-based feature extraction and the combination of CNN and transformer feature extraction for stereo matching.

\begin{table}[t]
  \centering
  
    \resizebox{0.48\textwidth}{!}{
    \begin{tabular}{lcc|ccc}
    \toprule
    Cost Volume &Dims& Channel& Flops&Params&EPE \\
    \midrule
    Difference &3D&-& 38.68G&2.40M& 1.02 \\
    \midrule
    Correlation &3D&-& 54.99G&4.01M& 0.81 \\
    \midrule
    Interlaced8&4D &8& 288.52G&2.83M& 0.70 \\
    \midrule
    Gwc8 &4D&  8  & 70.58G& 2.78M   & 0.72 \\
    Gwc16&4D& 16 & 166.92G&3.89M     & 0.66 \\
    Gwc24&4D&  24 &327.13G&5.75M     &0.63 \\
    Gwc32&4D&  32 &551.23G &8.34M    & 0.62 \\
    Gwc48&4D&  48 &1191.07G&15.73M     & \textbf{0.60}\\
    \midrule
    Cat24&4D  &  24 &328.97G&5.78M& 0.65\\
    Cat48&4D  &  48 & 1192.93G&15.76M&0.61\\
    Cat64&4D  & 64 & 2088.31G&26.09M&\textbf{0.60} \\
    \midrule
    G8-C16&4D&  24 & 328.96G&5.78M &0.62 \\
    G16-C24&4D&  40 & 841.05G&11.69M&\textbf{0.60} \\
    G32-C48&4D&  80 &3239.17G& 39.37M&\textbf{0.60} \\
    \bottomrule
    \end{tabular}%
    }
  \caption{\textbf{Ablation study on Cost Construction}. Gwc represents Group-wise correlation volume~\cite{gwcnet2019}. Cat stands for Concatenation volume~\cite{psmnet2018}. G8-C16, G16-C24, and G32-C48 combine Gwc volume and Cat volume~\cite{gwcnet2019}. Channel and Dims represent the channel and dimensions of the cost volume, respectively. \textbf{Bold}: Best.}
  \label{tab:ablation_cost}%
  \vspace{-0.7cm}
\end{table}%

\subsubsection{Cost Construction} 
In \autoref{tab:ablation_cost}, an ablation study on various cost volume strategies for stereo matching is presented. For these experiments, one-quarter of stereo image features are used to construct the cost volume. 
The study begins with simpler 3D cost volume methods: Difference~\cite{stereonet2018} and Correlation~\cite{wang2021fadnet++}, yielding higher EPE of 1.02 and 0.81, respectively, at lower computational costs. This suggests that while efficient, these methods may lack the nuanced disparity capture necessary for complex scenes. The Interlaced8 model, introduced by MobileStereoNet~\cite{shamsafar2022mobilestereonet}, achieves the same EPE comparable to the Gwc8 model. However, its computational expense is substantially higher, with a flop count of 288.52G, significantly larger than that of the Gwc8 model. The group-wise correlation and concatenation models demonstrate a clear trend: as the channel depth increases, the EPE improves, indicating improved disparity estimations through richer feature capture.
The combined volume (G8-C16) offers a more optimal balance between computational load and disparity estimation accuracy, which achieves an EPE of 0.62. G16-C24 and G32-C48 do not significantly improve EPE, despite a dramatic increase in computational load, especially for G32-C48, which demands 3239.17Gflops and has 39.37M parameters.
These results highlight the delicate balance between accuracy and computational efficiency in designing cost volumes for disparity estimation. While deeper and combined volumes reduce the EPE, the gains might be marginal compared to the significant increase in computational requirements, raising questions about the practicality of these approaches in resource-constrained environments.

\begin{table}[!t]
  \centering
  
  \resizebox{0.48\textwidth}{!}{
    \begin{tabular}{lc|ccc}
    \toprule
    Regression&Refinement&Flops & Params& EPE \\
    \midrule
    ArgMin &None&58.47G&2.69M &0.76 \\
    ArgMin &RGBRefine~\cite{xu2020aanet} & 117.85G&2.81M&0.72 \\
    ArgMin &Context &70.58G&2.78M&0.71  \\
    ArgMin &Context+RGBRefine~\cite{xu2020aanet} &129.95G&2.89M& 0.69 \\
    ArgMin &Context+DRNetRefine~\cite{xu2020aanet} & 129.88G& 2.89M&0.69 \\
    ArgMin &ConvGRU~\cite{raftstereo,xu2023iterative} & 3023.88G& 12.51M&\textbf{0.46} \\
    \bottomrule
    \end{tabular}
  }
  \caption{\textbf{Ablation study on Disparity Regression and Refinement.} ArgMin refers to Differentiable ArMin. Context stands for ContextUpasmple~\cite{raftstereo, xu2023iterative}. \textbf{Bold}: Best.}
  \label{tab:ablation_disparity}%
   \vspace{-0.5cm}
\end{table}%

\begin{figure*}[t]
\begin{center}
 \includegraphics[width=1\linewidth]{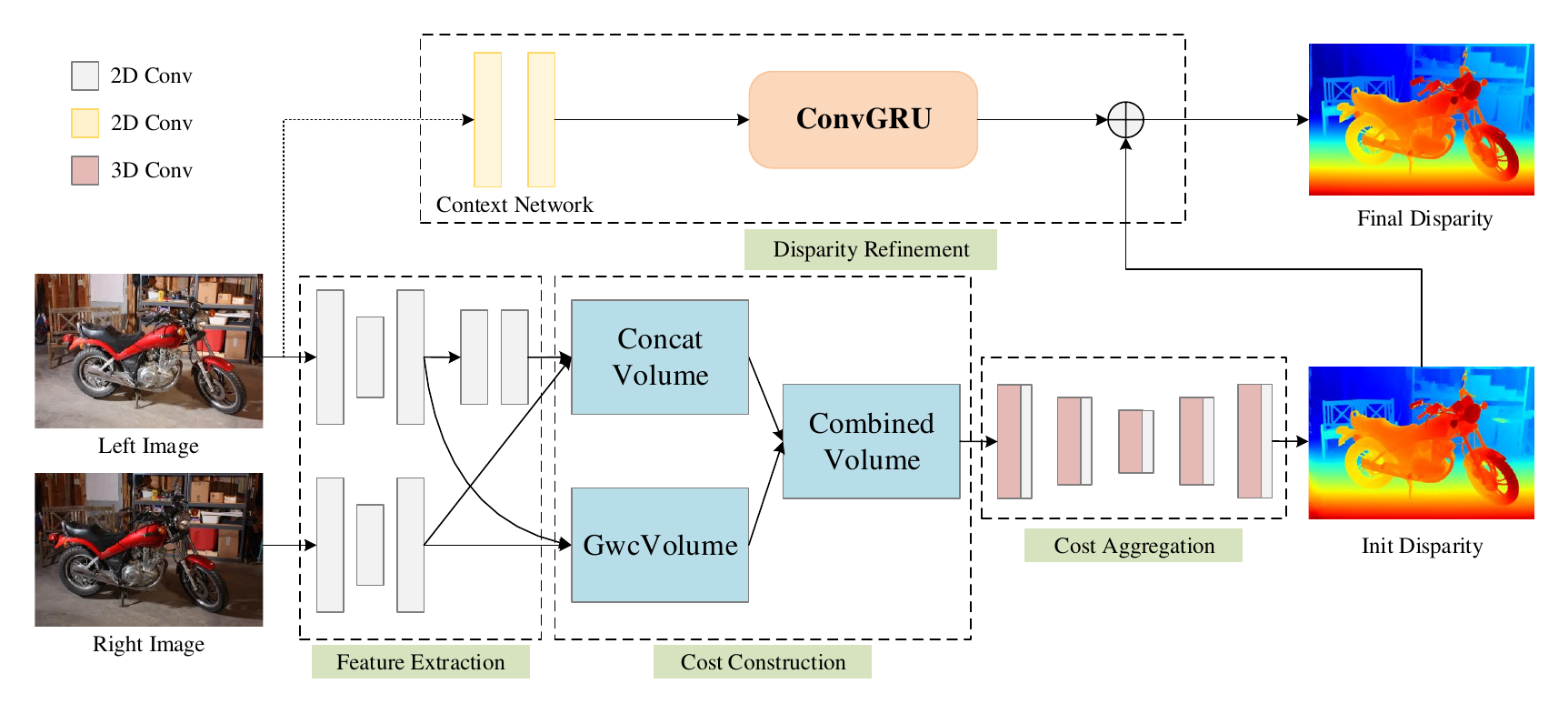}
\end{center}
\vspace{-0.8cm}
\caption{\textbf{Overview of our proposed StereoBase.} GwcVolume represents Group-wise correlation volume~\cite{gwcnet2019}.}
\vspace{-0.5cm}
\label{fig:stereobase}
\end{figure*}

\subsubsection{Disparity Regression and Refinement} 
The Differentiable ArgMin~\cite{gcnet,psmnet2018,gwcnet2019, cfnet2021,acvnet,gu2020cascade,ganet2019,shamsafar2022mobilestereonet, wang2021fadnet++} introduced by GCNet~\cite{gcnet}, calculates initial disparity by converting matching costs into probabilities via softmax and then computing a weighted sum of these probabilities across all disparity levels. 
As shown in~\autoref{tab:ablation_disparity}, various strategies show differing impacts on model performance in this ablation study on disparity refinement for the SceneFlow test datasets. Without refinement, the model has an EPE of 0.76. RGBRefine and Context methods slightly improve EPE to 0.72 and 0.71, respectively, with a modest increase in computational resources. Combining these methods further reduces EPE to 0.69, indicating marginal benefits from their integration. However, ConvGRU refinement substantially improves EPE to 0.46, albeit at a significant cost in computational complexity (3023.88 Gflops) and model size (12.51M). This highlights a trade-off between accuracy improvements and increased computational demands.


\section{A Strong Pipeline: StereoBase}



A strong baseline in deep stereo-matching research is critical for several key reasons. First, it serves as a vital reference point, enabling a clear assessment of new methods against an established standard. Second, a strong baseline allows for precise evaluation of the impact of specific changes, whether they are new data augmentation methods, different network architectures, or innovative disparity estimation techniques. This helps in isolating and understanding the contribution of each component to the overall performance. Additionally, a solid baseline ensures fair and meaningful comparisons across studies, providing a common ground for evaluating different research outcomes. This is crucial for maintaining consistency and validity in comparative analyses. 

\begin{table}[t]
    \centering
    
    \resizebox{0.5\textwidth}{!}{
    \begin{tabular}{l|c|cc|ccc}
     \toprule
     \multirow{2}{*}{Method} &SceneFlow & \multicolumn{2}{c|}{KITTI 2012} & \multicolumn{3}{c}{ KITTI 2015}  \\
     & EPE &3-noc & 3-all & D1-bg & D1-fg & D1-all\\
     \midrule
    PSMNet~\cite{psmnet2018} &1.09 &1.49 & 1.89  & 1.86 & 4.62 & 2.32 \\
    GwcNet~\cite{gwcnet2019}  &0.76 & 1.32 & 1.70  & 1.74 & 3.93 & 2.11 \\
    CFNet~\cite{cfnet2021}&1.04 &1.23&1.58&1.54&3.56&1.88\\
    
    
    CascadeStereo~\cite{gu2020cascade}&0.72&-&-& 1.59&4.03&2.00\\
    IGEV-Stereo~\cite{xu2023iterative}  &0.47 &1.12 & 1.44  &{1.38} &2.67& 1.59  \\
    GANet+ADL~\cite{ganetADL}&0.50&\textbf{0.98}&1.29& 1.38&2.38&1.55\\
    NMRF-Stereo~\cite{guan2024neural}&0.45&1.01&1.35 &\textbf{1.28}&3.13& 1.59\\
    Selective-IGEV~\cite{wang2024selective}&0.44&1.07&1.38& 1.33&2.61&1.55\\
    MoCha-Stereo~\cite{chen2024mocha} &0.41 &1.06&1.35 & 1.36&2.43&1.53\\
    \hline
    StereoBase (Ours) & \textbf{0.34}&1.00&\textbf{1.26}&\textbf{1.28}&\textbf{2.26}& \textbf{1.44} \\
    \bottomrule
    \end{tabular}
    }
    \caption{\textbf{Results on SceneFlow~\cite{sceneflow}, KITTI 2012~\cite{kitti2012}, and KITTI 2015~\cite{kitti2015} leaderboard.} \textbf{Bold}: Best.}
\label{tab:allleadboard}
\vspace{-0.35cm}
\end{table}

\begin{figure*}[t]
\begin{center}
 \includegraphics[width=1\linewidth]{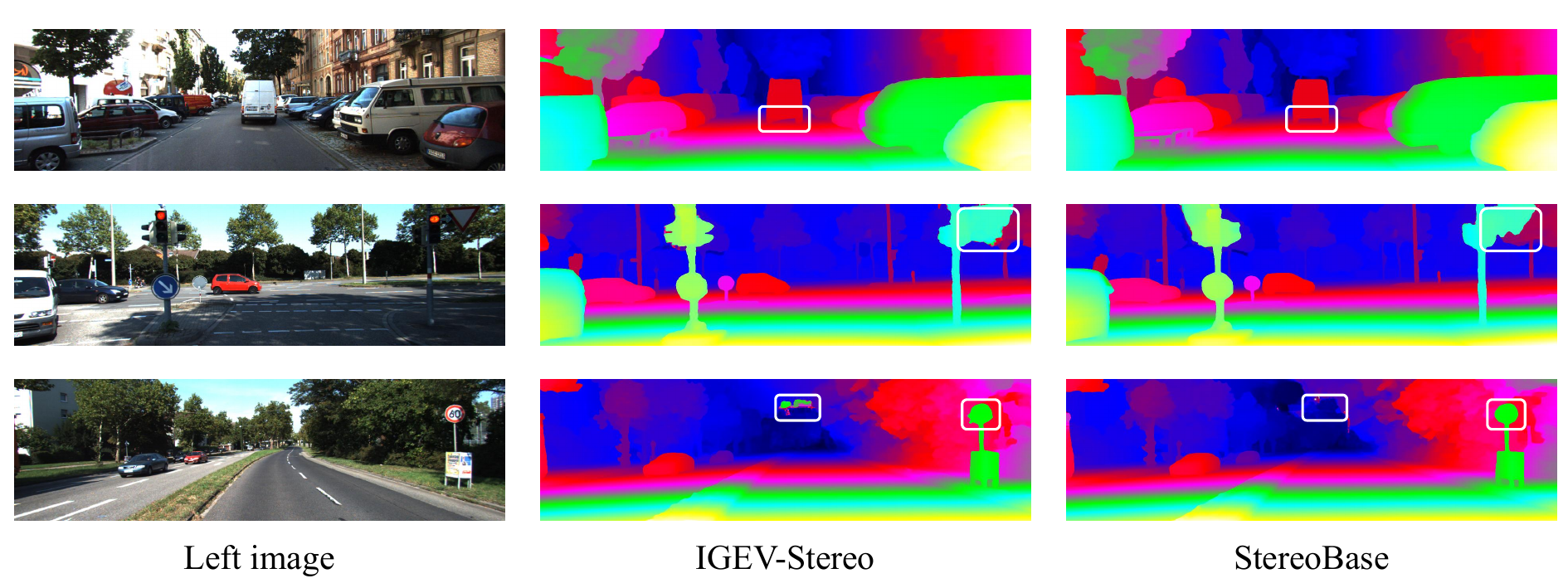}
\end{center}
\vspace{-0.5cm}
\caption{\textbf{Visualization results on KITTI2015 dataset.}}
\vspace{-0.5cm}
\label{fig:visualcomparsion}
\end{figure*}

\subsection{Pipeline}
In light of our comprehensive analysis, the goal of this section is to establish a strong baseline model that surpasses existing standards in performance. StereoBase embodies this objective. 
As shown in~\autoref{fig:stereobase}, given the left and the right images, the pre-trained MobileNetV2~\cite{rw2019timm} networks are used as our foundational backbone, extracting features at a reduced scale of 1/4th the original size to form the cost volume. The G8-C16 cost volume is utilized to achieve an optimal balance between computational load and disparity estimation accuracy. Hourglass networks~\cite{xu2023iterative} were implemented for cost aggregation, while convGRU~\cite{xu2023iterative} strategies were applied for the final disparity regression. 

\begin{table}[t]
    \centering
    
    \resizebox{0.48\textwidth}{!}{
    \begin{tabular}{l|cccc}
    \toprule
    Method & K12&  K15& Midd &ETH3D\\ 
      \midrule
STTR~\cite{STTR}$^\dagger$ & 49.72&40.26&OOM&38.89 \\
PSMNet~\cite{psmnet2018}$^\dagger$&30.51&32.15&33.53&18.02 \\
CFNet~\cite{cfnet2021}$^\dagger$&13.64&12.09&23.91&7.67  \\
AANet~\cite{xu2020aanet}$^\dagger$&7.23&7.72& 22.45&18.77 \\
M2D~\cite{shamsafar2022mobilestereonet}$^\dagger$& 18.34& 21.21&34.04  &13.89\\
M3D~\cite{shamsafar2022mobilestereonet}$^\dagger$& 25.14&22.44 & 29.32&13.71  \\
GwcNet~\cite{gwcnet2019}$^\dagger$&23.05 &25.19 &29.87 &14.54 \\
COEX~\cite{bangunharcana2021coex}$^\dagger$& 12.08& 11.01& 25.17& 11.43 \\
FADNet++~\cite{wang2021fadnet++}$^\dagger$& 11.31&13.23 & 24.07& 22.48 \\
CStereo~\cite{gu2020cascade}$^\dagger$&11.86 &12.06 &27.39 & 11.62  \\
IGEV~\cite{xu2023iterative}$^\dagger$&4.88 &\textbf{5.16}&\textbf{8.47}&3.53 \\
\midrule
StereoBase (Ours)& \textbf{4.85}&5.35&9.76&\textbf{3.12} \\
    \bottomrule
    \end{tabular}
    }
    \caption{\textbf{Cross-domain evaluation} on Middlebury, ETH3D, and KITTI all training sets. All methods are only trained on the Scene Flow dataset. Middlebury is tested on half-resolution. The model with $^\dagger$ indicates the implementation of OpenStereo. \textbf{Bold}: Best.}
\label{tab:crossdomain}
\vspace{-0.5cm}
\end{table}

\subsection{Comparison with State-of-the-art Methods}

\textbf{Implementation details} 
Training is driven by the smooth L1~\cite{psmnet2018} loss function. We utilize the AdamW optimizer in conjunction with the OneCycle learning rate scheduler.
For SceneFlow~\cite{sceneflow}, the total training epochs is set to 90. On the KITTI dataset, we finetune the pre-trained model using a combined dataset of KITTI2012~\cite{kitti2012} and 2015~\cite{kitti2015} training image pairs. The total epoch is 500. 

In our comprehensive evaluation, we benchmarked StereoBase against current state-of-the-art methods on SceneFlow~\cite{sceneflow}, KITTI2012~\cite{kitti2012}, and KITTI2015~\cite{kitti2015}.  
On the SceneFlow~\cite{sceneflow} test set, we achieve a new SOTA EPE of 0.34.
The quantitative comparisons, as summarized in \autoref{tab:allleadboard}, clearly illustrate the edge of StereoBase in handling complex stereo-matching scenarios with greater precision. Further, we submitted our results to the KITTI2012~\cite{kitti2012} and 2015~\cite{kitti2015} leaderboards, where StereoBase outperformed all published methods across all metrics. On KITTI2015~\cite{kitti2015}, our StereoBase outperforms IGEV~\cite{xu2023iterative} by 9.43\% on D1-all metric, respectively. In addition, we evaluate the generalization performance of StereoBase. As shown in \autoref{tab:crossdomain}, StereoBase exhibited exceptional performance in a zero-shot setting.
This evaluation further validates the adaptability and potential of StereoBase in handling diverse and challenging stereo vision tasks.


\section{Conclusion}
This paper introduces OpenStereo, a benchmark designed for deep stereo matching. Our initial endeavor involved re-implementing the most state-of-the-art methods within the OpenStereo framework. This comprehensive tool facilitates the extensive reevaluation of various aspects of stereo-matching methodologies. Drawing on the insights gained from our exhaustive ablation studies, we proposed StereoBase. Our StereoBase ranks 1\textsuperscript{st} on SceneFlow, KITTI 2015, 2012 (Reflective) among published methods and performs best across all metrics. In addition, StereoBase has strong cross-dataset generalization. StereoBase not only demonstrates the capabilities of our platform but also sets a new standard in the field for future research and development. Through OpenStereo and StereoBase, we aim to contribute a substantial and versatile resource to the stereo-matching community, fostering innovation and facilitating more effective and efficient research.



\bibliography{main}
\bibliographystyle{ieee_fullname}

\end{document}